\documentclass[letterpaper, 10 pt, conference]{ieeeconf}
\IEEEoverridecommandlockouts
\usepackage{amsmath}
\usepackage{mathtools}
\usepackage{cite}
\usepackage{amsmath,amssymb,amsfonts}
\usepackage[font=small, labelfont=bf]{caption}
\usepackage[ruled, lined, linesnumbered, commentsnumbered, longend, noend]{algorithm2e}
\usepackage{hyperref}
\usepackage{algorithmic}
\usepackage{graphicx}
\usepackage{textcomp}
\usepackage{xcolor}
\usepackage{breqn}
\usepackage{gensymb}
\usepackage{cuted}
\usepackage{capt-of}

\usepackage{pifont}
\newcommand{\cmark}{\ding{51}}
\newcommand{\xmark}{\textcolor{gray}{\ding{55}}}

\usepackage{fancyhdr} 
\def\BibTeX{{\rm B\kern-.05em{\sc i\kern-.025em b}\kern-.08em
    T\kern-.1667em\lower.7ex\hbox{E}\kern-.125emX}}

\definecolor{pink}{RGB}{255, 192, 203}

\definecolor{darkgreen}{RGB}{83, 199, 34}
\newcommand{\boldgreen}[1]{\textcolor{darkgreen}{\textbf{#1}}}

\title{\LARGE \bf Volumetric Mapping with Panoptic Refinement using\\Kernel Density Estimation for Mobile Robots}

\author{\large Khang Nguyen \hspace{45pt} Tuan Dang \hspace{45pt} Manfred Huber {\footnotesize \thanks{All authors are with the Learning and Adaptive Robotics Laboratory, Department of Computer Science and Engineering, University of Texas at Arlington, Arlington, TX 76013, USA. (emails: \href{mailto:khang.nguyen8@mavs.uta.edu}{\text{khang.nguyen8@mavs.uta.edu}}, \href{mailto:tuan.dang@uta.edu}{\text{tuan.dang@uta.edu}}, \href{mailto:huber@cse.uta.edu}{\text{huber@cse.uta.edu}})}}}

\begin{document}

\maketitle
\thispagestyle{empty}
\pagestyle{empty}

\begin{strip}
    \vspace{-53pt}
    \centering
    \includegraphics[width=1.00\linewidth]{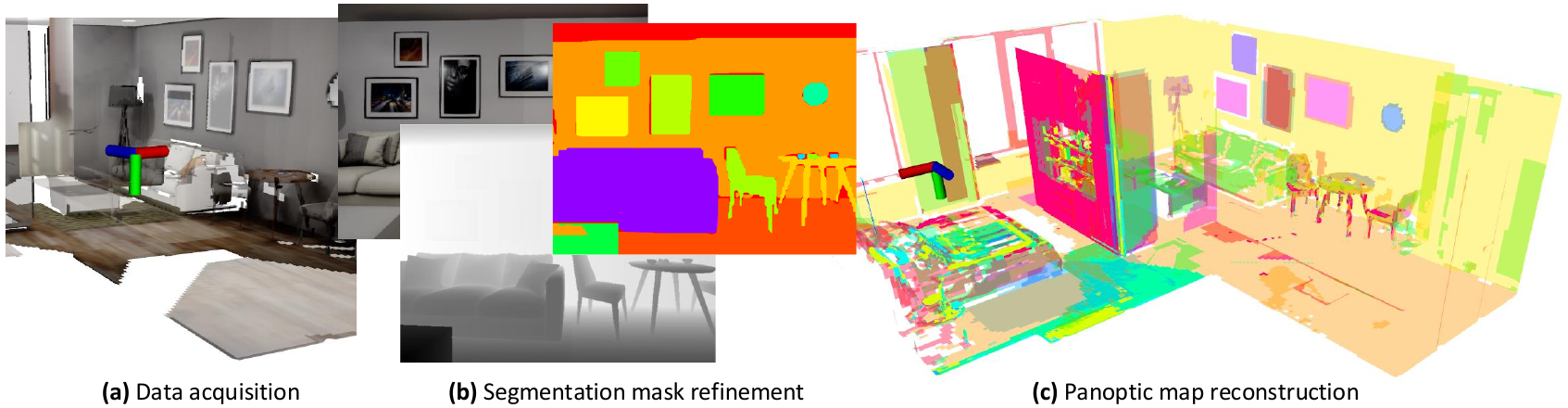}
    \vspace{-12pt}
    \captionof{figure}{\textbf{(a)} Indoor mobile robots operating in an environment with multiple objects \textbf{(b)} refines RGB-based segmentation masks using kernel density estimation via depth perception, and \textbf{(c)} rebuilds panoptic map with object instances using projective signed distance functions.}
    \vspace{-12pt}
    \label{fig:overview_method}
\end{strip}

\begin{abstract}
Reconstructing three-dimensional (3D) scenes with semantic understanding is vital in many robotic applications. Robots need to identify which objects, along with their positions and shapes, to manipulate them precisely with given tasks. Mobile robots, especially, usually use lightweight networks to segment objects on RGB images and then localize them via depth maps; however, they often encounter out-of-distribution scenarios where masks over-cover the objects. In this paper, we address the problem of panoptic segmentation quality in 3D scene reconstruction by refining segmentation errors using non-parametric statistical methods. To enhance mask precision, we map the predicted masks into a depth frame to estimate their distribution via kernel densities. The outliers in depth perception are then rejected without the need for additional parameters in an adaptive manner to out-of-distribution scenarios, followed by 3D reconstruction using projective signed distance functions (SDFs). We validate our method on a synthetic dataset, which shows improvements in both quantitative and qualitative results for panoptic mapping. Through real-world testing, the results furthermore show our method's capability to be deployed on a real-robot system. Our source code is available at: \href{https://github.com/mkhangg/refined_panoptic_mapping}{https://github.com/mkhangg/refined\_panoptic\_mapping}.
\end{abstract}

\section{Introduction}

Understanding and reconstructing 3D scenes are crucial in robotic perception and manipulation. SDF-based volumetric mapping methods are common in building 3D maps by integrating new observations from RGB-D images. To obtain the semantics of 3D maps, robots need to localize and identify each object in a scene, which is the first and foremost step toward scene understanding, where each voxel in the SDF-based map is assigned a label that can be predicted either from RGB-D images or point clouds. Nevertheless, robotic applications often rely on low-cost computations with information-rich RGB-D images to avoid the expense of computing and storing point clouds.

Traditional methods \cite{ren2012rgb, gupta2013perceptual, nguyen2024real} segment objects on RGB images using convolutional neural networks (CNNs) and look into depth maps to project 2D segments into 3D segments. Together with this, several methods \cite{le1995refining, freedman2005interactive, vicente2008graph, wang2020segmentation, grcic2023advantages} have been proposed to refine segments generated by CNNs using depth information. Depth-driven region growing method \cite{le1995refining} takes the similarity between objects' depth and connectivity criteria to produce accurate segmentation results, particularly for objects with distinct depth boundaries. Meanwhile, depth-assisted graph cut utilizes graph cut-based segmentation \cite{freedman2005interactive, vicente2008graph, wang2020segmentation, grcic2023advantages} to produce refined segmentation. However, depending solely on RGB images in the first place induces signification errors, especially on the border of segments if background colors are similar to the objects' colors.

Another approach is to fuse RGB and depth images using deep neural networks at the feature level; for example, the works \cite{li2016lstm, wang2016learning, park2017rdfnet, hazirbas2017fusenet, wang2018depth, yue2021two} design networks to explore the correlation between pixels of the same semantic and their corresponding depth under an assumption that pixels in the same segment should have similar depth and vice versa. Meanwhile, one of earlier works \cite{silberman2012indoor, couprie2013indoor, gupta2014learning} uses depth cues to support relations between objects to generate better segments. As RGB-D images are encoded in the network, it is difficult to detect minimal errors from the network output, and this largely depends on the nature of the training datasets. In other words, the main problem is that training and segmenting objects with RGB-D fusion does not guarantee the algorithm's performance will be adaptive under diverse real-world factors.

\begin{table*}[t]
    \vspace{4.5pt}
    \centering
    \resizebox{17.5cm}{!}{
    \begin{tabular}{c | c c c c c c c }    
        \hline
        \textbf{Features} & \textbf{KinectFusion} \cite{newcombe2011kinectfusion} & \textbf{Chisel} \cite{klingensmith2015chisel} & \textbf{Voxblox} \cite{oleynikova2017voxblox} & \textbf{Voxblox++} \cite{grinvald2019volumetric} & \textbf{Voxfield} \cite{pan2022voxfield} & \textbf{PanMap} \cite{schmid2022panoptic} & \textbf{Ours} \\
        \hline
        RGB-D-based panoptic perception & \xmark & \xmark & \xmark & \cmark & \xmark & \cmark & \cmark \\ 
        parametric-free segmentation refinement & \xmark & \xmark & \xmark & \xmark & \xmark & \xmark & \cmark \\ 
        SDF-based volumetric mapping & \cmark & \cmark & \cmark & \cmark & \cmark & \cmark & \cmark \\
        on-robot real-time performance & \cmark & \cmark & \cmark & \cmark & \cmark & \cmark & \cmark \\ 
        \hline
    \end{tabular}}
    \caption{Comparison of features across RGB-D volumetric mapping systems for indoor mobile robots.}
    \vspace{-17pt}
    \label{tab:features}
\end{table*}

To fill this gap, we propose a novel method to refine the over-covered segmentation masks generated by CNNs via Kernel Density Estimation (KDE) to eliminate the uncertainty in segmentation masks in a statistical manner, which differs from previous research where uncertainty is accumulated. The main advantage is that lightweight CNNs and KDE are suitable for resource-constrained embedded computers on robotic systems rather than point cloud-based segmentation networks. Additionally, with this proposed non-parametric method, we achieve a better result without fine-tuning models and hyperparameters for the pipeline to well-perform under various settings.

\section{Related Work}

\textbf{Panoptic Segmentation with RGB-D Perception}: Segmentation has recently gained popular attention in the robotics community thanks to its robustness in recognizing objects accurately at the pixel level. Earliest works, such as SemanticFusion \cite{mccormac2017semanticfusion}, Co-Fusion \cite{runz2017co}, and MaskFusion \cite{runz2018maskfusion}, fuse the semantics of objects onto simultaneous localization and mapping (SLAM) frameworks to build 3D maps; however, this method does not differentiate between objects of the same kind in the environment. To solve this, panoptic segmentation \cite{kirillov2019panoptic} then first brings the concept of semantic segmentation and instance segmentation together, where each image pixel is assigned an object label, benefiting robots to understand indoor entities distinctively. Leveraging panoptic segmentation, an incremental work \cite{nakajima2018fast} uses depth cues to segment objects but results in over-cover objects in complex scenes when objects are articulated with each other. Voxblox++ \cite{grinvald2019volumetric} and its subsequent PanMap \cite{schmid2022panoptic}, therefore, bridge the gap between these works by refining the mask and also using depth perception. Nevertheless, this geometric-based segmentation explicitly assumes that cutoff thresholds are provided when assigning 3D segments for each object instance. To ameliorate this, we proposed a depth-based segmentation refinement from the view of a non-parametric statistical approach, which enhances the adaptability of robots in various settings. 

\textbf{Outlier Rejection \& Mask Refinement}: Outlier removal is a crucial step in segmentation to ensure the robustness and precision of masks, particularly in dealing with learning models that generate noises in segments (\textit{i.e.,} at object boundaries). Classical methods that solely rely on the observation data are sensitive to noises without a good guess of hyperparameters, such as Random Sample Consensus (RANSAC) \cite{fischler1981random, yang2006robust} or distance-based outlier detection, such as Density-Based Spatial Clustering of Applications with Noise (DBSCAN) \cite{backlund2011density}. Recent approaches leverage the CNNs \cite{grcic2023advantages, wang2020segmentation} to encode RGB and depth images into separate encoders and use extracted features to correct the RGB-based segmentation error. Different approaches also fuse latent spaces of both RGB and depth as a single feature to generate correct segmentation masks. However, the primary limitation of these approaches is their dependency on training datasets, and their performance eventually subsides on out-of-distribution images and real-world scenarios. To avoid uncertainty in CNNs, we employ pre-trained 2D models to segment RGB images and refine these segments by applying KDE to each object's mask distribution on depth images.  

\textbf{SDF-Based Volumetric Mapping}: Modeling target objects' shapes and textures based on their colored images can be dated back to SDF from Curless and Levoy \cite{curless1996volumetric}. This has established the foundations for the recent emergence of robotics and graphics, especially volumetric mapping. Notable works in robotics using SDF can be categorized into object tracking \cite{schmidt2014dart, newcombe2015dynamicfusion, walsman2017dynamic} and volumetric modeling \cite{newcombe2011kinectfusion, oleynikova2016signed, oleynikova2017voxblox, grinvald2019volumetric, pan2022voxfield, schmid2022panoptic}, which are both essential components of mapping in the context of moving cameras. Signed Distance Functions \cite{oleynikova2016signed} and Voxblox \cite{oleynikova2017voxblox} first provide frameworks using SDF to model a 3D obstacle map for autonomous navigation based on prior works, such as KinectFusion \cite{newcombe2011kinectfusion} and Chisel \cite{klingensmith2015chisel}. 

Subsequently, Voblox++ \cite{grinvald2019volumetric} introduces functional scene understanding; meanwhile, Voxfield \cite{pan2022voxfield} is also developed with a more focus on accurate mapping but consumes lower computation costs. Most recently, PanMap \cite{schmid2022panoptic} presented a hierarchical semantic submap management. However, PanMap emphasizes the semantics of sub-maps more than object instances for ease of mapping management. Recognition inconsistency, therefore, affects points at the instance level, especially when an object occupies multiple voxels or parts of multiple recognized objects share one common voxel. Moreover, when segmentation masks are not well-refined via depth perception, these perception uncertainties can be increased in out-of-distribution scenes. To address this issue, in this work, we embed per-point semantics into the SDF-based volumetric mapping, where the voxels containing the points within the object of interest are updated incrementally, providing semantic consistency for the scene.

\section{Method Overview}
\label{sec:method_overview}

We first take an RGB-D image stream as shown in Fig. \ref{fig:overview_method}. With the RGB image, the segmentation model segments objects and non-object entities, such as walls, floors, and ceilings. The segmentation blobs are then projected using KDE on depth maps (Sec. \ref{sec:depth_outlier_rejection}) for depth point-based densities. Thus, the masks are refined based on density lines through depth perception in the previous step without additional hyperparameters (Sec. \ref{sec:segmentation_mask_refinement}). This mechanism effectively matches real-world intuition, in which the robots often encounter out-of-distribution scenes due to the effects of brightness and irregularity in objects' appearances and shapes. Iteratively, the scene is reconstructed via projective SDF, where the 3D points of objects of interest are updated over time (Sec. \ref{sec:projective_sdf_semantic_fusion}) until the robot stops its observation.

\begin{figure*}[t]
    \centering
    \includegraphics[width=1.00\linewidth]{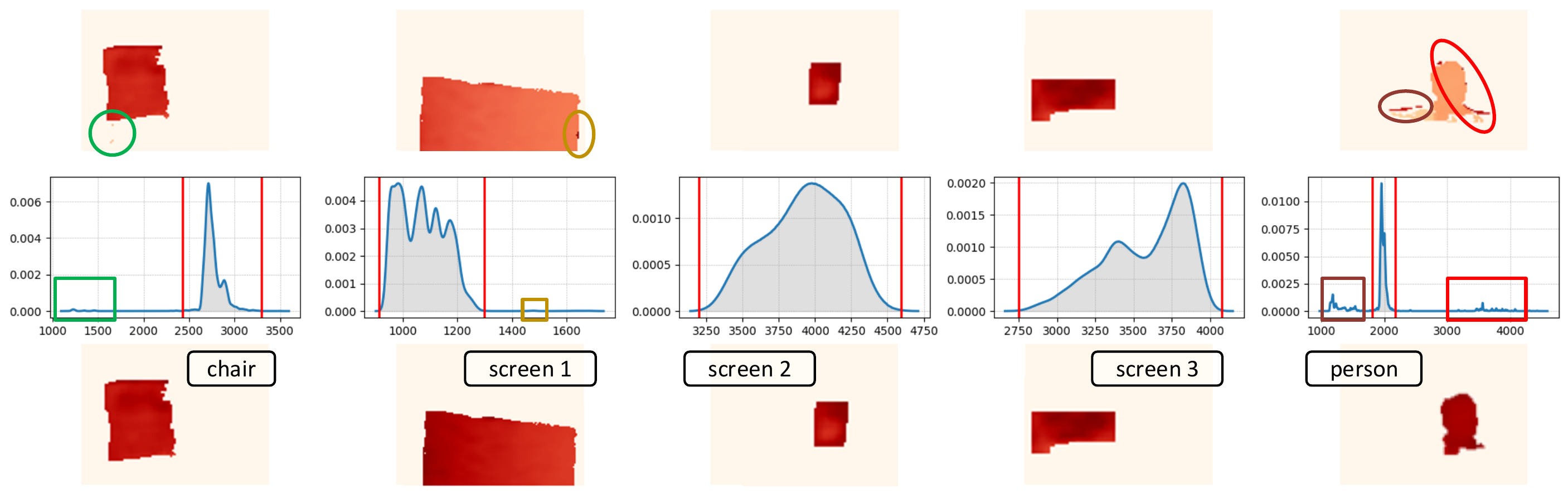}
    \vspace{-16pt}
    \caption{Depth maps of object instances containing depth outliers (\textit{top row}) due to the imperfection of segmentation models and their density estimations along depth perception (\textit{middle row}), and refined depth maps (\textit{bottom row}). The shaded depth values on the density lines in between vertical red cutoff lines are considered inliers; otherwise, Alg. \ref{alg:mask_refinement} rejects them as they appear to be outliers. The outliers are encoded by the same colors as Fig. \ref{fig:outlier_removal} along with corresponding objects presented in the scene.}
    \vspace{-13pt}
    \label{fig:kde_cutoffs}
\end{figure*}

\setlength{\textfloatsep}{0pt}
\begin{algorithm}[t]
    \caption{Mask Refinement via Depth Perception}
    \label{alg:mask_refinement}
    \begin{small}
        \DontPrintSemicolon
        \SetKwInOut{KwIn}{Input}
        \SetKwInOut{KwOut}{Output}
        \SetKwFunction{FMain}{RefineSegMaskViaDepth}
        \SetKwProg{Pn}{function}{}{}
        \KwIn{$\mathbf{M} \coloneqq$ binary masks of objects of interest\\
        $\mathcal{D} \coloneqq$ depth map}
        \KwOut{$\mathcal{M} \coloneqq$ refined masks for objects of interest}
        \Pn{\FMain{$\mathbf{M}$, $\mathcal{D}$}}{ 
            $\mathcal{M} = [\text{ }]$\\
            \For{$\mathbf{M}_{i} \in \mathbf{M}$}{
                $\mathcal{D}[\mathbf{M}_{i} = 0] = 0$\\
                $\mbox{\texttt{x\_kde},\text{ 
                }\texttt{y\_kde} = \texttt{FFTKDE}($\mathcal{D}$.\texttt{flatten}())}$\\
                ${peak} = \texttt{max}(\texttt{y\_kde})$\\
                $\texttt{left\_id} = \texttt{find\_id}(\texttt{y\_kde}[:peak] < \texttt{1e-6})$\\
                $\texttt{right\_id} = \texttt{find\_id}(\texttt{y\_kde}[peak:] < \texttt{1e-6})$\\
                $\texttt{low\_cutoff} = \texttt{x\_kde}[\texttt{left\_id}]$\\
                $\texttt{high\_cutoff} = \texttt{x\_kde}[\texttt{right\_id}]$\\
                $\mathcal{D}\left[d < \texttt{low\_cutoff} \textbf{ or } d > \texttt{high\_cutoff}\right] = 0$\\
                $\mathbf{M}_{i}[\mathcal{D} = 0] = 0$\\
                $\mathcal{M}.\texttt{append}(\mathbf{M}_{i})$
            }
            \KwRet{$\mathcal{M}$}
        }
    \end{small}
\end{algorithm}

Our approach additionally offers a parametric-free depth outlier rejection for segmentation mask refinement, as depicted in Table \ref{tab:features}. This improves the quality of SDF-based panoptic mapping and makes robots, particularly mobile indoor robots, adaptive and versatile in out-of-distribution scenarios, which has not been well-addressed in previous works of the same category.   

\section{Methodology}
\label{sec:methodology}

Mobile robots usually produce uncertainties in their perception, which are best seen via depth maps, where depth pixels cannot be interpolated from the previous frame, resulting in holes on depth maps. Plus, in 3D semantic perception, RGB-based segmentation models add uncertainty when recognizing objects and mapping their occupancies in the real world. To alleviate this, we consider addressing two uncertainties: (1) depth map hole-filing and (2) refining segmentation masks in 3D space.

\subsection{Holes Filling on Depth Images}

To fill possible holes that occurred on depth maps, $\mathcal{D}$, we interpolate the missing values from its $k \times k$ neighboring pixels. Otherwise, the missing value remains as 0 if all of their $g^2$ neighbors are empty, as follows:
\begin{equation}
    \small
    \mathcal{D}(i, j) = \left\{
    \begin{array}{ll}
        0 \text{, if } \mathcal{D}(i \pm k, j \pm l) = 0 \\
         \displaystyle \sum_{i=u-k}^{u+k} \sum_{j=v-k}^{v+k} \mathcal{D}(i, j) \cdot G(u, v)
    \end{array} \right. 
    \label{eq:hole_filling}
\end{equation}
where $0 \leq k, l \leq g$, $(u, v)$ represents the image coordinates of pixels, $G(u, v)$ is the 2D Gaussian kernel, size of $g \times g$, centered on the $(u, v)$ pixel. 

\begin{figure}[t]
    \centering
    \includegraphics[width=1.00\linewidth]{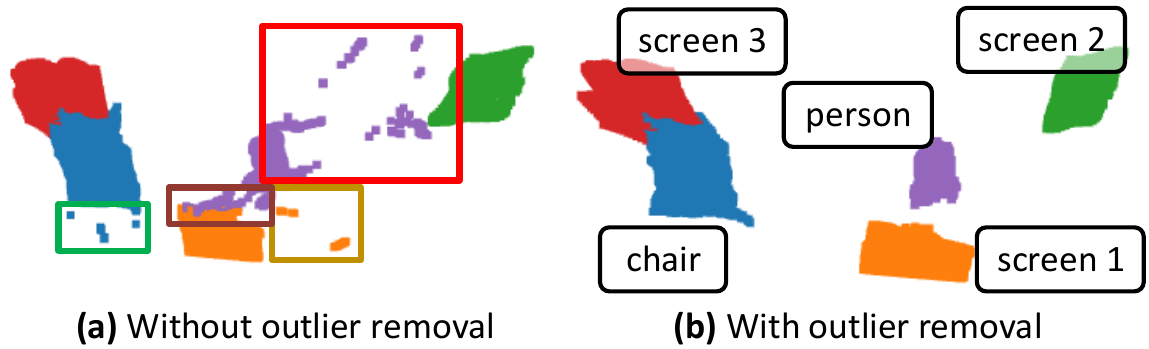}
    \caption{The scene of multiple objects with outliers boxed in red (\textit{left}) and the scene without outliers after applying Alg. \ref{alg:mask_refinement} (\textit{right}).}
    \label{fig:outlier_removal}
\end{figure}

\subsection{Depth Outlier Rejection}
\label{sec:depth_outlier_rejection}

Segmentation imperfection also occurs in objects' boundary pixels, leading to inaccurate depth perception when aligning RGB and depth frames. In practice, this is typically ignored by manually truncating depth pixels that exceed a defined threshold. Another approach to overcome this is taking the top-down view, where depth errors are projected on the table surface and spatially compensated within the manipulation process. To solve this, we apply the density function  $\hat{f}$ at any given point $d$ to non-parametrically reject depth outliers as follows:
\begin{equation}
    \small
    \hat{f}(d) = \frac{1}{mH}\sum^{m}_{i=1} \mathcal{G}\left(\frac{d - d_{i}}{H}\right)
    \label{eq:kde}
\end{equation}
where $d_{i}$ is the depth value from $\mathcal{D}(\cdot, \cdot)$, $H$ represents the optimal bandwidth obtained from the ISJ algorithm for the 1D Gaussian kernel, $\mathcal{G}$, and $m$ indicates the number of depth values on the segmented object.

Re-organizing Eq. \ref{eq:kde} in terms of equidistant 1D grid points $\{\textbf{g}_{j}\} \in \left[\textbf{g}_{1}, \textbf{g}_{M}\right]$ covering all $\textbf{p}^{\mathcal{K}_{t}}_{i}$, and grid counts $\{c_{j}\}$ to represent the number of $\textbf{p}^{\mathcal{K}_{t}}_{i}$'s that are near $\textbf{g}_{j}$, for $j = 1, 2, ..., M$ with $M \neq m$, we obtain:
\begin{equation}
    \small
    \widetilde{f_{d, \textbf{g}_j}} \coloneqq \tilde{f}(g_{j}) = \frac{1}{mH}\sum^{M}_{i=1}  c_{i} \cdot \mathcal{G}\left(\frac{\textbf{g}_{j} - \textbf{g}_{i}}{H}\right) \approx \hat{f}(d)
    \label{eq:binning_kde}
\end{equation}

\begin{figure*}[t]
    \centering
    \includegraphics[width=1.00\linewidth]{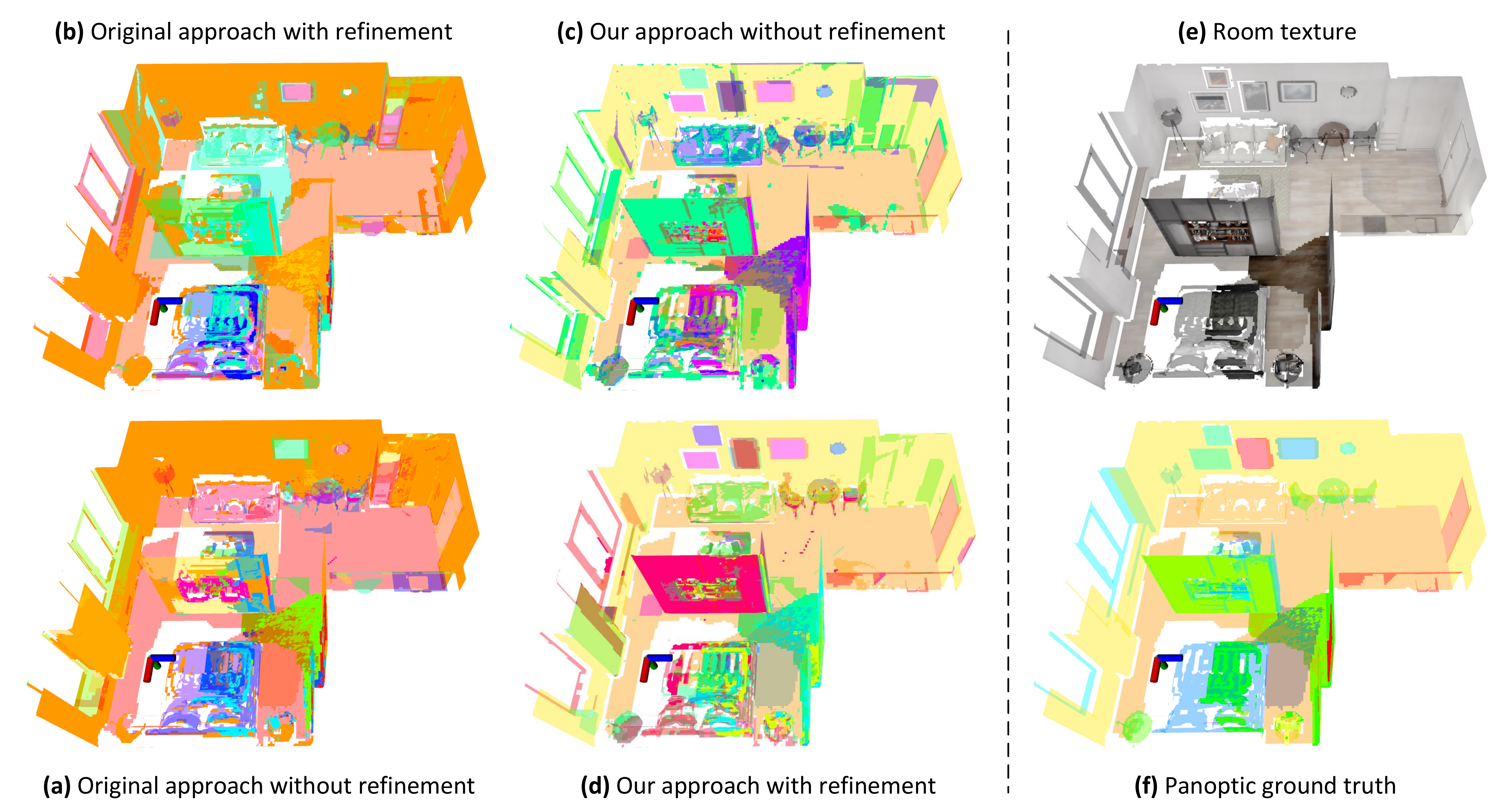}
    \vspace{-12pt}
    \caption{Qualitative results on the \texttt{flat} dataset of \textbf{(a)} the original panoptic mapping approach, \textbf{(b)} the original approach coupled with mask refinement, \textbf{(c)} our approach without mask refinement, and \textbf{(d)} our approach with mask refinement. The room texture and its panoptic segmentation ground truth are retrieved based on RGB images and annotation masks provided by the original framework \cite{schmid2022panoptic}. Note that the robot frame indicating its pose is simplified and represented as the RGB mesh frame in each reconstructed map.}
    \vspace{-15pt}
    \label{fig:qualitative_results}
\end{figure*}

With the FFT's time complexity of $\mathcal{O}(M\log M)$, Eq. \ref{eq:binning_kde} is then translated in the form of convolution as:
\begin{equation}
    \small
    \widetilde{f_{d, \textbf{g}_j}} = \sum^{M-1}_{i=-(M-1)} c_{j-i} \textbf{k}_{i} \text{ with } \textbf{k}_{i} = \frac{1}{m} \cdot \mathcal{G}\left(\frac{\textbf{g}_{M} - \textbf{g}_{1}}{H} \cdot i\right)
    \label{eq:fft_kde}
\end{equation}

\subsection{Segmentation Mask Refinement}
\label{sec:segmentation_mask_refinement}

Alg. \ref{alg:mask_refinement} illustrates the segmentation mask refinement process via depth perception, given the set of RGB-D binary segmentation masks and the depth map. For each object, its depth values are selected according to its binary mask. Hence, applying Eq. \ref{eq:fft_kde}, we obtain the density function along with depth values. The maximum peak in the density line is then identified; meanwhile, the cutoff values on its two tails are determined when the density goes to 0 -- indicating the disconnection between the object and its fragments in the depth axis, as shown in Fig. \ref{fig:kde_cutoffs}. The depth map is rectified on regions where the depth pixels sub-ceed the lower cutoff and exceed the upper cutoff, followed by the mask refinement via non-zero regions of the depth map. Iterating through objects of interest, Alg. \ref{alg:mask_refinement} returns their refined binary masks with outlier rejection via depth knowledge without requiring predefined thresholds.  

As an example provided in Fig. \ref{fig:outlier_removal}a, the outliers induced by segmentation masks are visible in the form of point clouds. By applying the refined binary masks returned by Alg. \ref{alg:mask_refinement}, the outliers are also removed in the corresponding point clouds of instances, as shown in Fig. \ref{fig:outlier_removal}b.

\subsection{Integration of Projective SDFs \& Semantic Perception}
\label{sec:projective_sdf_semantic_fusion}

Comprehending the semantics of indoor environments with a mobile robot entails dynamically updating the occupancy in the real world. Embedded with semantic knowledge along with the building process, the robot is also ready for other manipulation tasks besides navigating and exploring such environments. Therefore, to do this in the discretized voxel-like world, our objective is to incrementally update in 3D space via SDF \cite{curless1996volumetric} with semantic knowledge in a recursive manner, as follows:
\begin{equation*}
    \small
    D_{t + 1}(v, \hat{\textbf{p}}) = \frac{W_{t}(v, \hat{\textbf{p}}) D_{t}(v, \hat{\textbf{p}}) + w_{t + 1}(v, \hat{\textbf{p}}) d_{t + 1}(v, \hat{\textbf{p}})}{W_{t}(v, \hat{\textbf{p}}) + w_{t + 1}(v, \hat{\textbf{p}})}
\end{equation*}
\begin{equation}
    \small
    W_{t}(v, \hat{\textbf{p}}) = \sum_{i} w_{t - 1}^{(i)}(v, \hat{\textbf{p}}) + w_{t}(v, \hat{\textbf{p}})
    \label{eq:tsdf_semantic}
\end{equation}
\begin{equation*}
    \small
    d_{t + 1}(v, \hat{\textbf{p}}) = \left\{
    \begin{array}{ll}
        || \hat{\textbf{p}} - v ||_{2}\text{, if } \hat{\textbf{p}} \in v \\
        - || \hat{\textbf{p}} - v ||_{2}\text{, if } \hat{\textbf{p}} \not\in v \\
    \end{array} \right. 
\end{equation*}
where $v$ is the voxel in 3D, $\hat{\textbf{p}}$ is the 3D point within in the object of interest, $d_{i}(\cdot, \cdot)$ represents SDFs with their corresponding weights, $w_{i}(\cdot, \cdot)$, from refined masked RGB images, and $D_{i}(\cdot, \cdot)$ indicates the cumulative SDF with $W_{i}(\cdot, \cdot)$ as the corresponding cumulative weight function.

Using Eq. \ref{eq:tsdf_semantic}, the voxels containing the points within the object of interest are updated recursively, eventually providing the robot with the object's occupancy voxels.

\section{Evaluation on \texttt{flat} Dataset}
\label{sec:evaluations}

\subsection{Dataset \& Evaluation Metrics}

We evaluate Alg. \ref{alg:mask_refinement} on the \texttt{flat} dataset to verify our approach's performance in both quantitative and qualitative results against the most recent state-of-the-art method \cite{schmid2022panoptic}. The \texttt{flat} dataset consists of synthetic RGB-D images rendered in Unreal Engine 4 (UE4) and includes ground truth mesh, per-image panoptic annotations, and the camera poses at each RGB-D frame. In terms of quantitative results, we compute the mask intersection over union (IOU) between the ground truth annotations and the predicted masks provided by the segmentation model. Meanwhile, in terms of qualitative results, we compare the results after performing volumetric mapping across the approaches. In addition, we also reconstruct the scene with texture and panoptic masks to better compare the 3D reconstruction qualities.

\begin{table}[h]
    \centering
    \vspace{-4pt}
    \resizebox{8.6cm}{!}{
    \begin{tabular}{c | c c c }    
        \hline
        \textbf{Approaches} & \textbf{Mask IOU} & \textbf{Changes} \\
        \hline
        \textbf{(a)} PanMap without refinement & 16.5150 & -- \\ 
        \textbf{(b)} PanMap with refinement & 26.2283 & \boldgreen{+9.7133 $\uparrow$} \\
        \textbf{(c)} Our approach without refinement & 79.8860 & \boldgreen{+53.6577 $\uparrow$} \\ 
        \textbf{(d)} Our approach with refinement & 90.6077 & \boldgreen{+10.7217 $\uparrow$} \\
        \hline
    \end{tabular}}
    \caption{Quantitative results in mask IOU percentage and changes on the \texttt{flat} dataset between \textbf{(a)} the original approach without mask refinement, \textbf{(b)} with mask refinement, \textbf{(c)} our approach without mask refinement, and \textbf{(d)} with mask refinement.}
    \vspace{-15pt}
    \label{tab:quantitative_results}
\end{table}

\subsection{Quantitative Results}
\label{sec:quantitative_results}
The quantitative comparisons between our approach and the panoptic mapping technique \cite{schmid2022panoptic} are shown in Table \ref{tab:quantitative_results}. For the original approach, the mask IOU is at 16.5150; however, after refining their predicted masks, the mask IOU is elevated by approximately 10 percent to 26.2283. 

Since the semantic consistency at the point level is not prioritized in the original approach, we retrain the segmentation model \cite{yolov8} with similar objects and produce segmentation masks for the same sequence; however, we do not apply the mask refinement process initially. We thus achieve the mask IOU of 79.8860, which significantly improves from Detectron used in the original approach. Therefore, to see how mask refinement's effectiveness on new sets of segmentation images, we proceed with performing mask refinement for each mask in the set; the result then reaches 90.6077 in terms of mask IOU, again improving by roughly 10 percent from when not applying mask refinement.

Overall, we observe a slight improvement when applying mask refinement to the raw set of segmentation masks. This implies the crucial role of refinement based on depth maps, which strongly correlate to the objects' spatial occupancy and shapes in the 3D world rather than on 2D images.

\subsection{Qualitative Results}
We continue to compare the quality of SDF-based volumetric mapping on the same robot moving sequence between the original and our approaches, as shown in Fig. \ref{fig:qualitative_results}. Also, for each approach, we compare the results with and without the mask refinement step to see how crucial it is qualitatively.

\begin{figure}[t]
    \centering
    \includegraphics[width=1.00\linewidth]{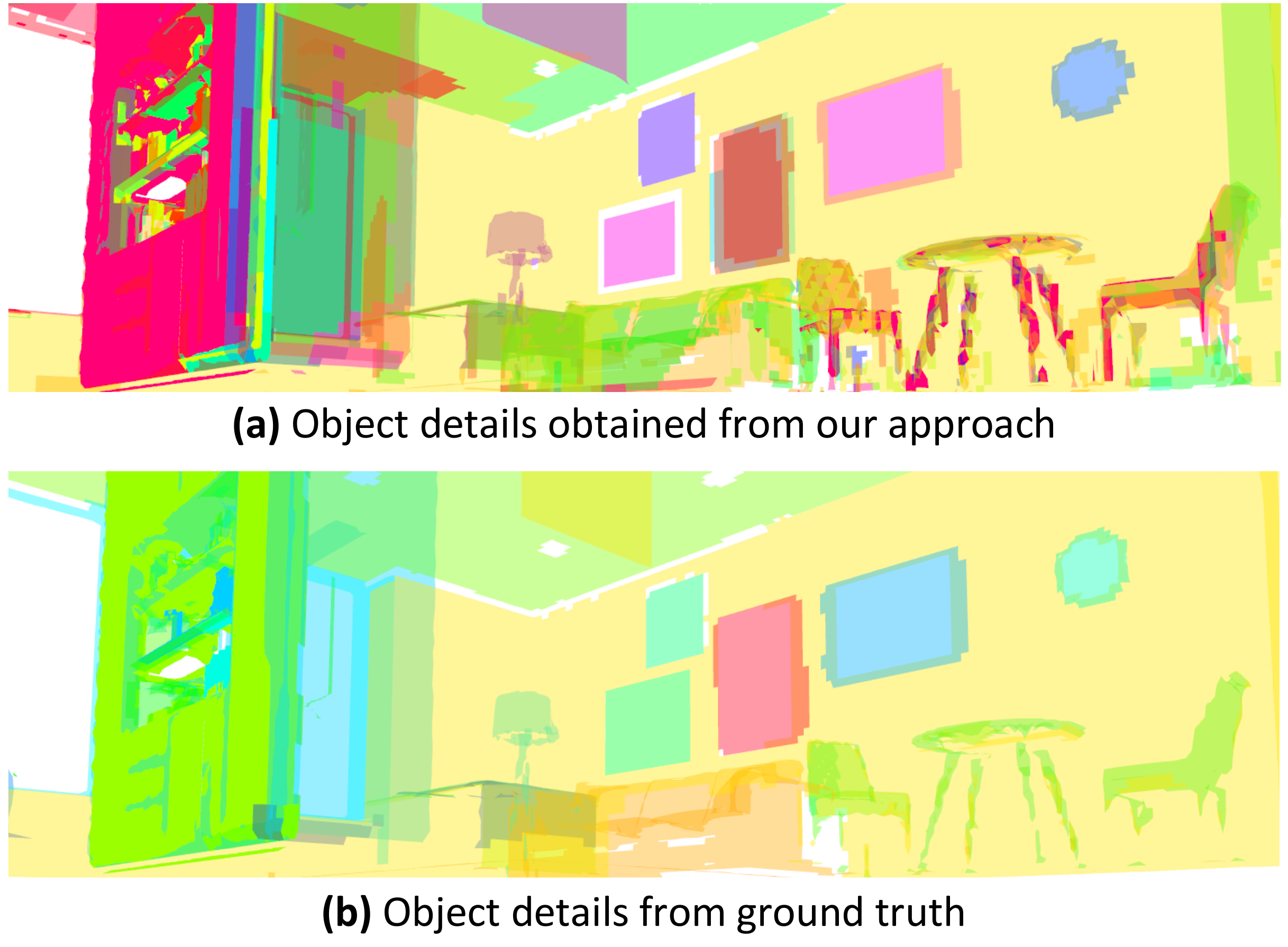}
    \caption{Comparisons of object detail reconstruction quality between \textbf{(a)} our approach with mask refinement and \textbf{(b)} from ground truth.}
    \label{fig:object_details}
    \vspace{7pt}
\end{figure}

\begin{figure*}[t]
    \centering
    \includegraphics[width=1.00\linewidth]{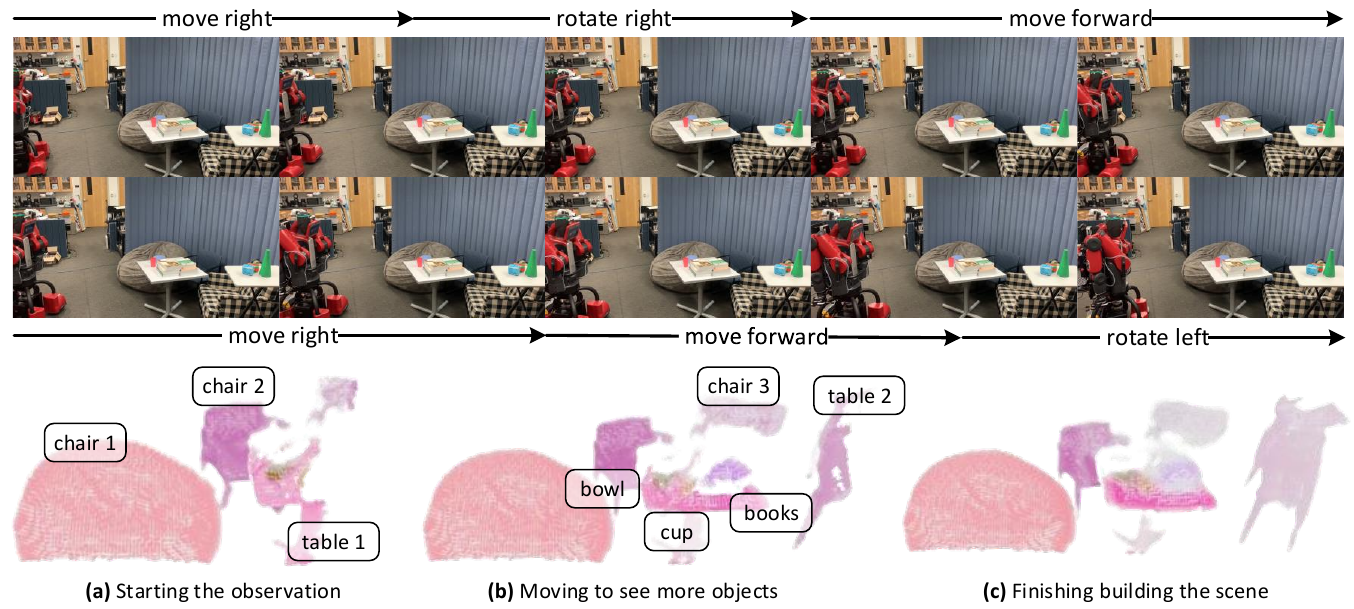}
    \vspace{-12pt}
    \caption{The refined volumetric mapping process on the Baxter robot: \textbf{(a)} starts to observe a part of the scene, \textbf{(b)} iteratively updating the scene by moving in the lab's free space, including translations and rotations, and \textbf{(c)} finishing building the observed scene.}
    \vspace{-15pt}
    \label{fig:experiments}
\end{figure*}

\subsubsection{Original Approach without Mask Refinement} As shown in Fig. \ref{fig:qualitative_results}a, the raw panoptic mapping approach mainly focuses on guaranteeing the scene's hierarchical map management, where the semantic consistency of sub-maps is prioritized over the completeness and accuracies of the segmentation. However, this approach remains pitfalls when the recognition inconsistency occurs at the instance level, resulting in point-level inconsistency, especially in circumstances where an object straddles multiple submaps or parts of multiple objects occupy one submap simultaneously. These cases are brittle, especially when segmentation masks are not well-refined via depth perception.

\subsubsection{Original Approach with Mask Refinement} Therefore, to mitigate this error, we perform the refinement process on each segmentation mask. As shown in Fig. \ref{fig:qualitative_results}b, the other picture on the wall is partially detected and reconstructed, which has been entirely omitted in Fig. \ref{fig:qualitative_results}a. Nevertheless, the qualitative result is far off compared to the ground truth in Fig. \ref{fig:qualitative_results}f. The reason for this is due to the imperfection of naive segmentation inputs, inducing false detections at instance levels.

\subsubsection{Our Approach without Mask Refinement}
To resolve this problem, we use the retrained segmentation model as mentioned in Sec. \ref{sec:quantitative_results} and perform re-segmentation on RGB images, thus reconstructing the scene with new segmentation masks. The result in Fig. \ref{fig:qualitative_results}c shows that the approach is able to recognize more objects compared to those in Fig. \ref{fig:qualitative_results}b. However, the outlier artifacts are presented due to the lack of segmentation mask refinement. 

\subsubsection{Our Approach with Mask Refinement}
Combining the mask refinement process with the approach in Fig. \ref{fig:qualitative_results}, we effectively remove small-sized outliers and achieve a ``cleaner" map at the end. Compared to the original result in Fig. \ref{fig:qualitative_results}a, we are able to segment and reconstruct objects of interest in the room, such as tables, chairs, sofas, clocks, wall pictures, etc, with respect to the room texture in Fig. \ref{fig:qualitative_results}e and the panoptic segmentation ground truth in Fig. \ref{fig:qualitative_results}f. In this approach, these object details (Fig. \ref{fig:object_details}a) are also well-constructed compared to the ground truth (Fig. \ref{fig:object_details}b).

\section{Real-Robot Experiments}

To test the adaptability of the mask refinement procedure in real-world settings, we employ our proposed method on a real-robot system with a setup scene of everyday objects.

\subsection{Software \& Hardware Setup} 
We build our software on distributed computers where all components are synchronized using a Robotic Operating System (ROS). One node in ROS is responsible for acquiring the RGB-D image steam from the depth camera and pre-processing raw data before broadcasting them into the ROS network. The pre-processing step includes segmenting RGB images and refining their segmentation masks using our mentioned method above. Meanwhile, another node performs SDF-based volumetric mapping when receiving refined segments from the data acquisition ROS node. Note that each node runs different computers, and each RGB-D image time-stamp synchronizes raw data. 

\subsection{Real-World Performance of Proposed Method}
We set up the scene of indoor objects containing chairs and tables of various types with stacks of books and a plastic cup on the table, as shown in the top right of Fig. \ref{fig:experiments}. The robot is operated to walk around the scene arbitrarily with the support of omnidirectional wheels. In Fig. \ref{fig:experiments}a, the robot bootstraps the pipeline and starts observing the setup scene, and the scene is incrementally built as the robot moves. Initially, the bean bag, chair, part of the table, and cup appeared (Fig. \ref{fig:experiments}a). Then, the other chair, the rest of the table, and the stack of books are built (Fig. \ref{fig:experiments}b). Lastly, the robot tries to complete the scene by observing more points of the tables, which are also incrementally updated into the existing reconstructed one (Fig. \ref{fig:experiments}c). The walls and floors are colored in grey for ease of visualizing per-object refinements.

\subsection{Spatial Mask Refinement \& Semantic Consistency}
Also, as shown in the bottom row of Fig. \ref{fig:experiments}, each reconstructed objects are spatially well-refined using mask refinement. These refinements are visible at corners of objects, such as chair legs and table legs, and in between objects' boundaries. Likewise, per-point semantic consistency is well-preserved via projective SDF-based volumetric mapping. Together with the refinements, these qualitative results underline the adaptability of our proposed method in real-world indoor settings and on a real-robot system.

\subsection{Demonstration}
The demonstration video shows the performance of our proposed method on the Baxter mobile robot is available at \url{https://youtu.be/u214kCms27M}.

\section{Conclusions}

In this paper, we present a parametric-free mask refinement process for projective SDF-based volumetric mapping, which improves the quality of 3D scene reconstruction with panoptic understanding. The segmentation-induced outliers appearing when masking onto objects' point clouds are statistically removed by applying KDE along the depth perception, whereas the discontinuity on density lines is rectified automatically without additional hyperparameters. After the segmentation mask refinement process, the point of each object instance is updated via projective SDF, which, together with the mask precision, guarantees point-level semantic consistency when volumetrically building the panoptic map. To verify our proposed method's precision, we conduct evaluations on the synthetic \texttt{flat} dataset and achieve better results in both quantitative and qualitative manners. The results of experiments on the Baxter robot with an Intel RealSense D435i RGB-D camera also demonstrate that our method is adaptive and feasible in real-world indoor environments.

\bibliographystyle{IEEEtran}
\bibliography{IEEEabrv, 09_references}

\end{document}